\documentclass[journal, onecolumn,12pt]{IEEEtran}

\usepackage[margin=20mm]{geometry}
\usepackage{booktabs}
\usepackage{epsfig}
\usepackage{graphicx}
\usepackage{amssymb}
\usepackage{amsfonts}
\usepackage{multirow}
\usepackage{algorithmic}
\usepackage{amsmath}
\usepackage{amsthm}
\usepackage[linesnumbered,ruled,vlined]{algorithm2e}
\usepackage{mathtools}
\usepackage[dvipsnames,table]{xcolor}
\usepackage{subfigure}
\usepackage{tikz}
\usepackage{paralist}
\usepackage{color}
\usepackage{tabu}
\usepackage{rotating}
\usepackage{pifont}
\usepackage{cite}
\usepackage{threeparttable}
\usepackage{moresize}
\usepackage{cancel}
\usepackage{pgf-pie}
\usepackage{adjustbox}
\usepackage{authblk}
\usepackage{url}

\usepackage[colorlinks=true,linkcolor=black,anchorcolor=black,citecolor=black,filecolor=black,menucolor=black,runcolor=black,urlcolor=blue]{hyperref}

\def\hl{\setlength{\fboxsep}{1.0pt}\colorbox[rgb]{0.85,0.85,0.85}}

\newtheorem*{example*}{Example}

\renewcommand{\vec}{\mathbf}

\usepackage{authblk}

\title{\vspace{-1.5cm}\textbf{Competition on Dynamic Optimization Problems Generated by Generalized Moving Peaks Benchmark (GMPB)}}

\author{Danial~Yazdani,
 Michalis~Mavrovouniotis,
 Changhe~Li,
 Guoyu~Chen,
 Wenjian~Luo,
 Mohammad~Nabi~Omidvar,
 J\"{u}rgen~Branke,
Shengxiang~Yang, and
Xin~Yao

\thanks{Danial Yazdani is with the Business Intelligence Team, WINC, Australia (email: danial.yazdani@gmail.com).}
\thanks{Michalis~Mavrovouniotis is with Cyprus University of Technology, Lemesos, Cyprus (e-mail: michalis.mavrovouniotis@eratosthenes.org.cy).}
\thanks{Changhe~Li is with the School of Artificial Intelligence, Anhui University of Sciences \& Technology, Hefei, China (e-mail: changhe.lw@gmail.com).}
\thanks{Guoyu~Chen is with the School of Artificial Intelligence, Anhui University of Sciences \& Technology, Hefei, China (e-mail:  chenguoyumail@163.com).}
\thanks{Wenjian~Luo is with the Guangdong Provincial Key Laboratory of Novel Security Intelligence Technologies, School of Computer Science and Technology, Harbin Institute of Technology and Peng Cheng Laboratory, Shenzhen 518055, China (e-mail: luowenjian@hit.edu.cn).}
\thanks{Mohammad Nabi Omidvar is with the School of Computing, University of Leeds, and Leeds University Business School, Leeds LS2 9JT, United Kingdom (e-mail: m.n.omidvar@leeds.ac.uk).}
\thanks{J\"{u}rgen Branke is with the Operational Research and Management Sciences Group at Warwick Business school, University of Warwick, Coventry CV4 7AL, United Kingdom (e-mail: Juergen.Branke@wbs.ac.uk}
\thanks{Shengxiang~Yang is with the Center for Computational Intelligence (CCI), School of Computer Science and Informatics, De Montfort University, Leicester LE1 9BH, United Kingdom (e-mail: syang@dmu.ac.uk).}
\thanks{Xin Yao is with the Department of Computing and Decision Sciences, Lingnan University, Hong Kong, China. 
He is also with the CERCIA, School of Computer Science, Birmingham B15 2TT, United Kingdom (e-mail: xinyao@LN.edu.hk).} 
}

\begin{document}

\maketitle

\begin{abstract} 
The Generalized Moving Peaks Benchmark (GMPB)~\cite{yazdani2020benchmarking} is a tool for generating continuous dynamic optimization problem instances with controllable dynamic and morphological characteristics. 
GMPB has been used in recent Competitions on Dynamic Optimization at prestigious conferences, such as the IEEE Congress on Evolutionary Computation (CEC). 
This dynamic benchmark generator can create a wide variety of landscapes, ranging from simple unimodal to highly complex multimodal configurations, and from symmetric to asymmetric forms. It also supports diverse surface textures, from smooth to highly irregular, and can generate varying levels of variable interaction and conditioning. This document provides an overview of GMPB, emphasizing how its parameters can be adjusted to produce landscapes with customizable characteristics. The MATLAB implementation of GMPB is available on the \href{https://github.com/EDOLAB-platform/EDOLAB-MATLAB}{\textcolor{blue}{\underline{EDOLAB platform}}}~\cite{peng2023evolutionary}. 
\end{abstract}

 

\section{Introduction}
Search spaces of many real-world optimization problems are dynamic in terms of the objective function, the number of variables, and/or constraints~\cite{nguyen2012evolutionary,nguyen2011thesis}. 
Optimization problems that change over time and need to be solved online by an optimization method are referred to as dynamic optimization problems~(DOPs)~\cite{yazdani2021DOPsurveyPartA}. 
To solve DOPs, algorithms not only need to find desirable solutions but also to react to the environmental changes in order to quickly find a new solution when the previous one becomes suboptimal~\cite{yazdani2018thesis}.

To comprehensively evaluate the effectiveness of algorithms designed for DOPs, a suitable benchmark generator is crucial.
Commonly used and well-known benchmark generators in this domain use the idea of having several components that form the landscape. 
In most existing DOP benchmarks, the width, height, and location of these components change over time~\cite{cruz2011optimization}.  
One of the benchmark generators that is designed based on this idea is the Moving Peaks Benchmark (MPB)~\cite{branke1999memory}, which is the most widely used synthetic problem in the DOP field~\cite{cruz2011optimization,yazdani2018thesis,nguyen2011thesis}. 
Each component in MPB is formed by usually a simple peak whose basin of attraction is determined using a $\max(\cdot)$ function. 

Despite the popularity of the traditional MPB, landscapes generated by this benchmark consist of components that are smooth/regular, symmetric, unimodal, separable~\cite{yazdani2019scaling}, and easy-to-optimize, which may not be the case in many real-world problems. 
In~\cite{yazdani2020benchmarking}, a \emph{generalized} MPB (GMPB) is proposed which can generate components with a variety of characteristics that can range from unimodal to highly multimodal, from symmetric to highly asymmetric, from smooth to highly irregular, and with different variable interaction degrees and condition numbers. 
GMPB is a benchmark generator with fully controllable features that helps researchers to analyze DOP algorithms and investigate their effectiveness in facing a variety of different problem characteristics.

\section{Generalized Moving Peaks Benchmark~\cite{yazdani2020benchmarking}}
\label{sec:GMPB}

The baseline function of GMPB is\footnote{Herein, we focus on fully non-separable problem instances which are constructed by Eq.~\eqref{eq:irGMPB}.
Those readers who are interested in partially/fully separable problems (i.e., composition based functions), refer to~\cite{yazdani2020benchmarking}.}:
\begin{align}
\label{eq:irGMPB}
 f^{(t)}(\vec{x})= \max_{k\in\{1,\dots,m\}}\left\{ h_k^{(t)} - \sqrt{\mathbb{T}\left(\left(\vec{x}-\vec{c}_{k}^{(t)}\right)^\top{\mathbf{R}_{k}^{(t)}}^\top,k\right) ~\mathbf{W}_{k}^{(t)}~ \mathbb{T}\left(\mathbf{R}_{k}^{(t)}\left(\vec{x}-\vec{c}_{k}^{(t)}\right),k\right)} \right\},
\end{align}
where $\mathbb{T}(\vec{y},k):\mathbb{R}^{d}\mapsto\mathbb{R}^{d}$ is calculated as: 
\begin{align}
\label{eq:ir}
    \mathbb{T}\left(y_{j},k\right)=
    \begin{dcases}
    \exp{\left(\log(y_{j})+\tau^{(t)}_{k}\left(\sin{(\eta_{k,1}^{(t)}\log(y_{j}))}+\sin{(\eta_{k,2}^{(t)}\log(y_{j}))} \right)\right)} & \text{if   } y_{j}>0 \\
    0 & \text{if   }y_{j}=0\\
    -\exp{\left(\log(|y_{j}|)+\tau^{(t)}_{k}\left(\sin{(\eta_{k,3}^{(t)}\log(|y_{j}|))}+\sin{(\eta_{k,4}^{(t)}\log(|y_{j}|))} \right)\right)} & \text{if   } y_{j}<0  
\end{dcases}
\end{align}
where $\vec{x}$ is a solution in $d$-dimensional space,  $m$ is the number of components, $\mathbf{R}_{k}^{(t)}$ is the rotation matrix of $k$th component in the $t$th environment, $\mathbf{W}_{k}^{(t)}$ is a $d \times d$ diagonal matrix whose diagonal elements show the width of $k$th component  in different dimensions,  
$y_{j}$ is $j$th dimension of $\mathbf{y}$, and $\eta_{k,l\in\{1,2,3,4\}}^{(t)}$ and $\tau^{(t)}_{k}$  are irregularity parameters of the $k$th component.

For each component $k$, the rotation matrix $\mathbf{R}_{k}$ is obtained by rotating the projection of $\vec{x}$ onto all $x_p$-$x_q$ planes by a given angle $\theta_{k}$. 
The total number of unique planes which will be rotated is ${d \choose 2} = \frac{d(d-1)}{2}$. 
For rotating each  $x_p$-$x_q$ plane by a certain angle ($\theta_{k}$), a \emph{Givens rotation matrix} $\mathbf{G}_{(p,q)}$ must be constructed.
To this end, first, $\mathbf{G}_{(p,q)}$ is initialized to an identity matrix $\mathbf{I}_{d \times d}$; then, four elements of $\mathbf{G}_{(p,q)}$ are altered as: 
\begin{align}
\mathbf{G}_{(p,q)}(p,p)&= \cos\left(\theta_{k}^{(t)}\right),\\
\mathbf{G}_{(p,q)}(q,q)&= \cos\left(\theta_{k}^{(t)}\right),\\
\mathbf{G}_{(p,q)}(p,q)&=  -\sin\left(\theta_{k}^{(t)}\right),\\
\mathbf{G}_{(p,q)}(q,p)&=  \sin\left(\theta_{k}^{(t)}\right),
\end{align}
where $\mathbf{G}_{(p,q)}(\mathfrak{i},\mathfrak{j})$ is the element at $\mathfrak{i}$th row and $\mathfrak{j}$th column of $\mathbf{G}_{(p,q)}$.
$\mathbf{R}_{k}$ in the $t$th environment is calculated by:
\begin{align}
\label{eq:allrotation}
 \mathbf{R}_{k}^{(t)}= \left( \prod_{(p,q)\in \mathcal{P}} \mathbf{G}_{(p,q)}\right)   \mathbf{R}_{k}^{(t-1)},
\end{align}
where $\mathcal{P}$ contains all unique pairs of dimensions defining all possible planes in a $d$-dimensional space. 
The order of the multiplications of the \emph{Givens rotation matrices} is random.
The reason behind using~\eqref{eq:allrotation} for calculating $\mathbf{R}$ is that we aim to have control on the rotation matrix based on an angle severity $\tilde{\theta}$. 
Note that the initial $\mathbf{R}^{(0)}_{k}$ for problem instances with rotation property is obtained by using  the Gram-Schmidt orthogonalization method on a matrix with normally distributed entries. 

For each component $k$, the height, width vector, center, angle, and irregularity parameters change from one environment to the next according to the following update rules:
\begin{align}
\vec{c}_{k}^{(t+1)}&=\vec{c}_{k}^{(t)} + \tilde{s} \frac{\vec{r}}{\|\vec{r}\|},  \label{eq:center} \\
h_{k}^{(t +1)}&=h_{k}^{(t)}  + \tilde{h} \, \mathcal{N}(0,1), \label{eq:height} \\
w_{k,j}^{(t +1)}&=w_{k,j}^{(t)}  + \tilde{w} \, \mathcal{N}(0,1), j\in\{1,2,\cdots,d\},\label{eq:width} \\
\theta_{k}^{(t +1)}&= \theta_{k}^{(t)}+\tilde{\theta} \, \mathcal{N}(0,1),\label{eq:angle}\\
\eta_{k,l}^{(t +1)}&= \eta_{k,l}^{(t)}+\tilde{\eta} \, \mathcal{N}(0,1), l\in\{1,2,3,4\},\label{eq:tau}\\
\tau_{k}^{(t +1)}&= \tau_{k}^{(t)}+\tilde{\tau} \, \mathcal{N}(0,1),\label{eq:eta}
\end{align}
where $\mathcal{N}(0,1)$ is a random number drawn from a Gaussian distribution with mean 0 and variance 1, $\mathbf{c}_{k}$ shows the vector of center position of the $k$th component, $\vec{r}$ is a $d$-dimensional vector of random numbers generated by $\mathcal{N}(0,1)$,  $\|\vec{r}\|$ is the Euclidean length (i.e., $l_2$-norm) of $\vec{r}$,  $\frac{\vec{r}}{\|\vec{r}\|}$ generates a unit vector with a random direction, $\tilde{h}$, $\tilde{w}$, $\tilde{s}$, $\tilde{\theta}$, $\tilde{\eta}$, and $\tilde{\tau}$ are height, width, shift, angle, and two irregularity parameters' change severity values, respectively, $w_{k,j}$ shows the width of the $k$th component in the $j$th dimension, and $h_{k}$ and $\theta_{k}$ show the height and angle of the $k$th component, respectively.

Outputs of equations~\eqref{eq:center}~to~\eqref{eq:eta} are bounded as follows: $h_{k}\in[h_{\mathrm{min}},h_\mathrm{max}]$, $w_{k}\in[w_\mathrm{min},w_\mathrm{max}]^{d}$, $\vec{c}_{k}\in[Lb,Ub]^{d}$, $\tau\in[\tau_\mathrm{min},\tau_\mathrm{max}]$, $\eta_{1,2,3,4}\in[\eta_\mathrm{min},\eta_\mathrm{max}]$, and $\theta_{k}\in[\theta_\mathrm{min},\theta_\mathrm{max}]$, where $Lb$ and $Ub$ are maximum and minimum problem space bounds. 
For keeping the above mentioned values in their bounds, a \emph{Reflect} method is utilized. Assume $a^{(t+1)}=a^{(t)}+b$ represents one of the equations~\eqref{eq:center}~to~\eqref{eq:eta}. 
The output based on the reflect method is:
\begin{gather}
\label{chap5:eq:survivalTime}
a^{(t+1)}=
\begin{dcases}
a^{(t)}+b & \text{if   }a^{(t)}+b\in[a_\mathrm{min},a_\mathrm{max}] \\
2\times a_\mathrm{min}-a^{(t)}-b & \text{if   }a^{(t)}+b<a_\mathrm{min} \\
2\times a_\mathrm{max}-a^{(t)}-b & \text{if   }a^{(t)}+b>a_\mathrm{max} 
\end{dcases}
\end{gather}

\subsection{Problem characteristics}

In this section, we describe the main characteristics of the components and the landscapes generated by GMPB.
Note that in the 2-dimensional examples provided in this part, the values of width (2-dimensional vector), irregularity parameters ($\tau$ and four values for $\eta$), and rotation status (1 is rotated and 0 otherwise) are shown in the title of figures. 

\subsubsection{Component characteristics}

\begin{figure}[tp!]
\centering
\begin{tabular}{cc}
     \subfigure[{\scriptsize }]{\includegraphics[width=0.45\linewidth]{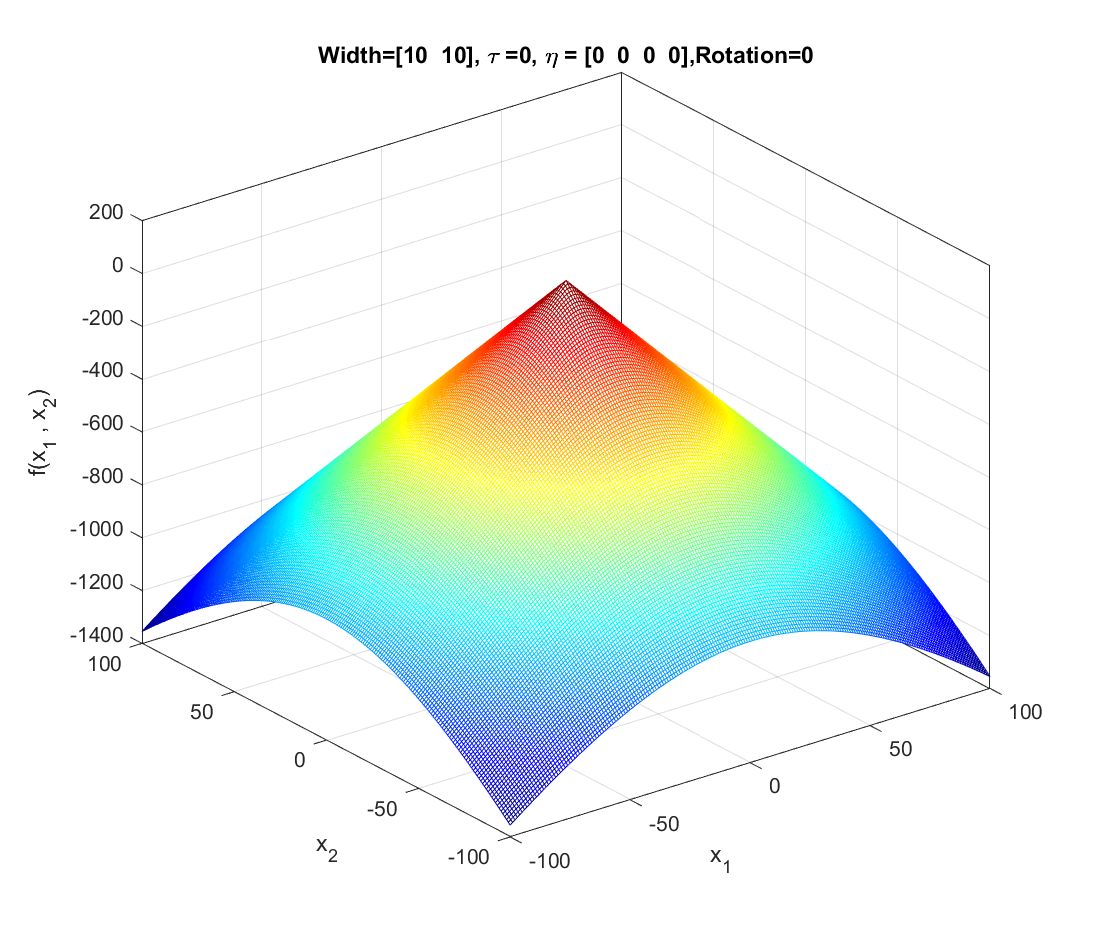}\label{fig:Cmponent:cone:surf}}
&
    \subfigure[{\scriptsize }]{\includegraphics[width=0.45\linewidth]{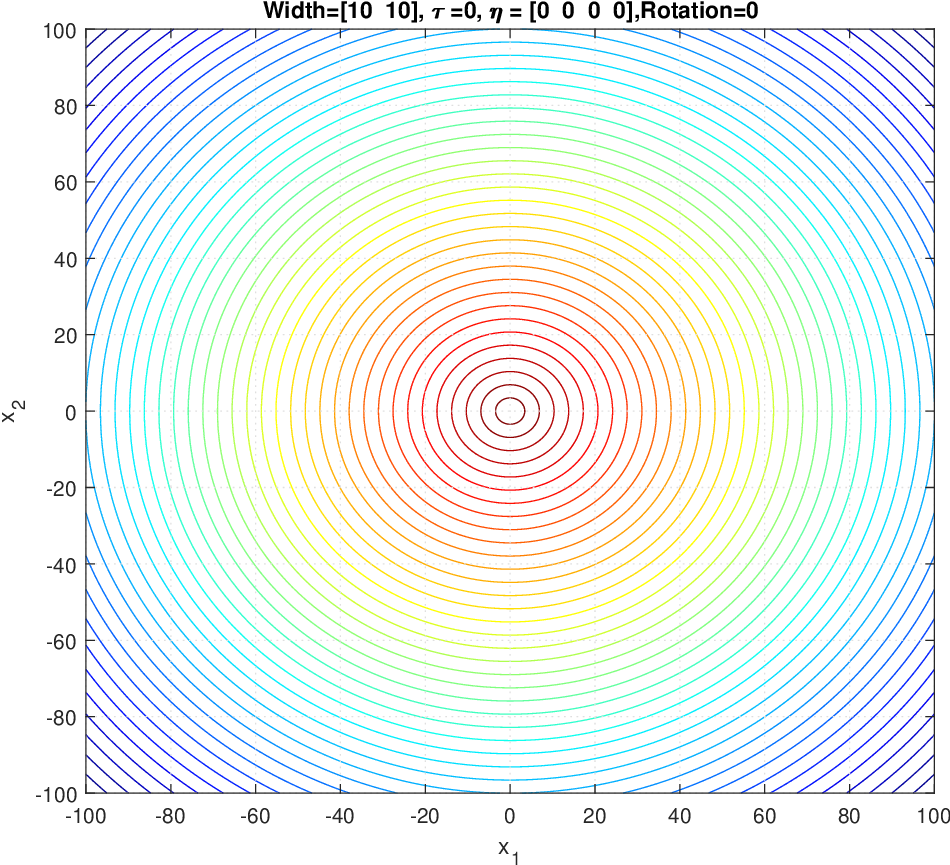}\label{fig:Cmponent:cone:contour}}
\end{tabular}
\caption{A conical component (regular, smooth, and unimodal) generated by \eqref{eq:irGMPB} whose width values are identical, $\mathbf{R}=\mathbf{I}$, and $\tau$ and $\eta$ are set to zero.}
\label{fig:Cmponent:cone}
\end{figure}

\begin{figure}[tp!]
\centering
\begin{tabular}{cc}
     \subfigure[]{\includegraphics[width=0.45\linewidth]{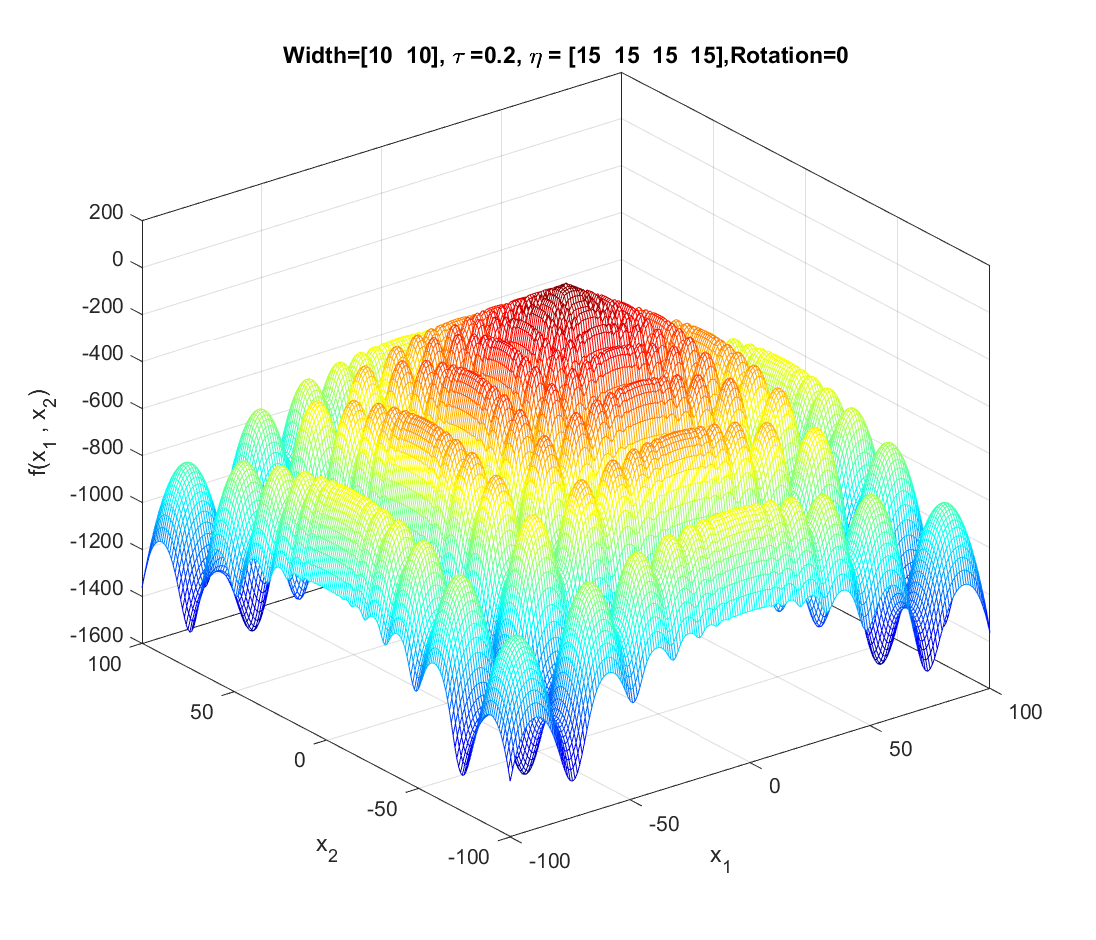}\label{fig:Cmponent:irregular:surf}}
&
    \subfigure[{\scriptsize }]{\includegraphics[width=0.45\linewidth]{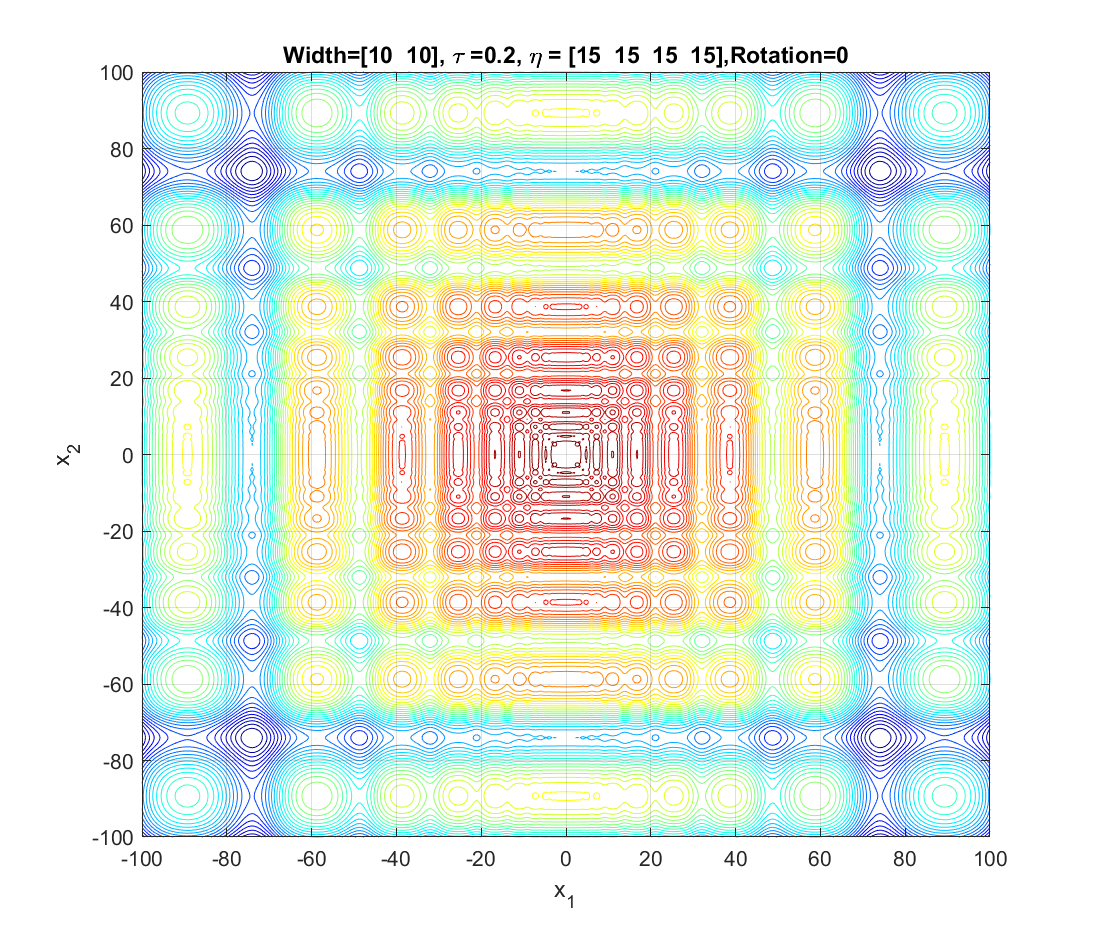}\label{fig:Cmponent:irregular:contour}}
    \\
         \subfigure[]{\includegraphics[width=0.45\linewidth]{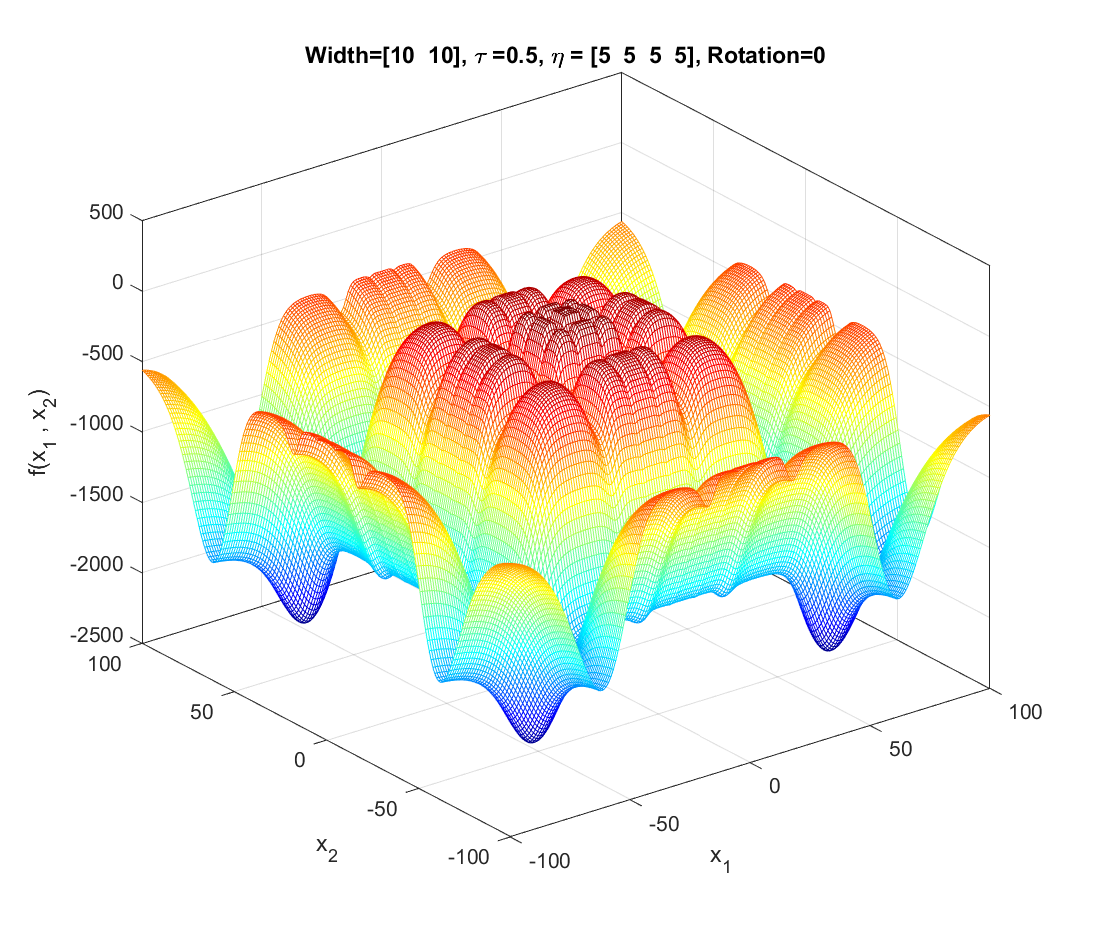}\label{fig:Cmponent:irregular(Challenging):surf}}
&
    \subfigure[{\scriptsize }]{\includegraphics[width=0.45\linewidth]{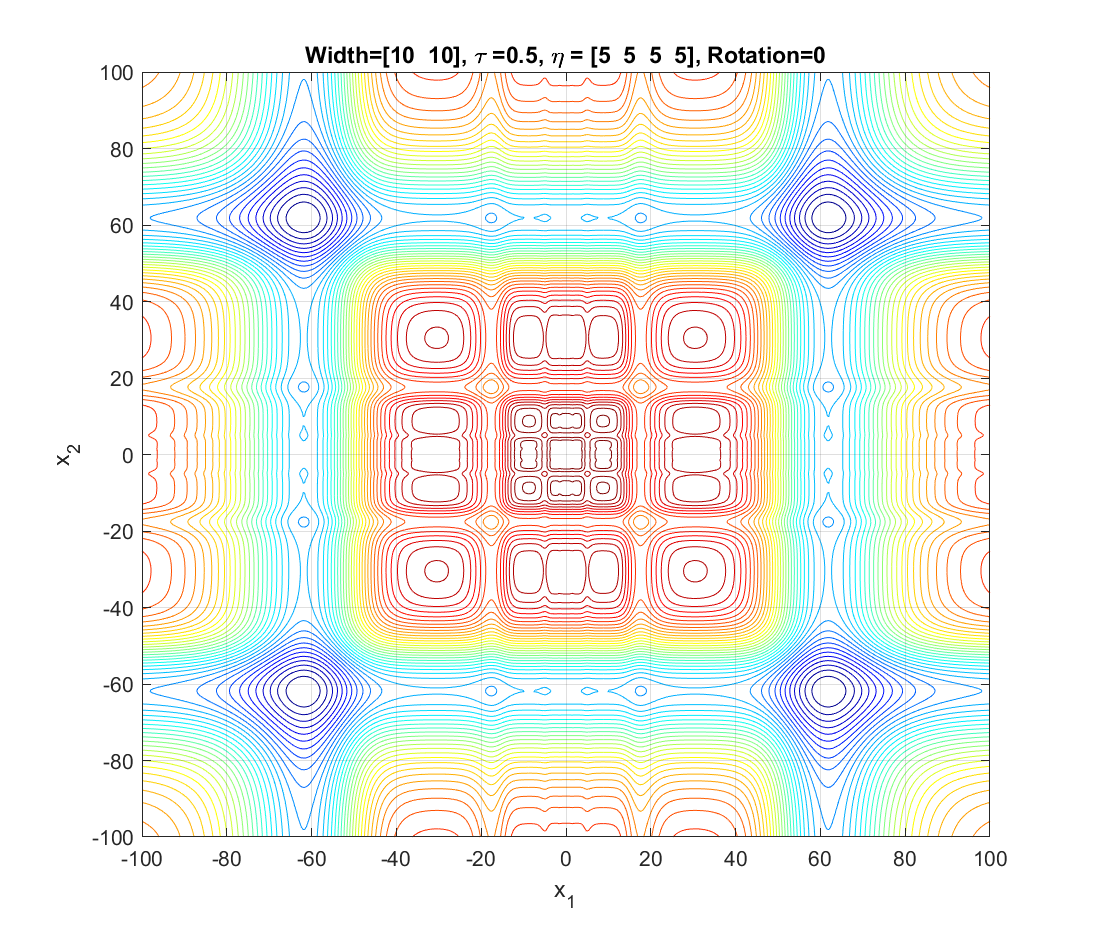}\label{fig:Cmponent:irregular(Challenging):contour}}
   \\
         \subfigure[]{\includegraphics[width=0.45\linewidth]{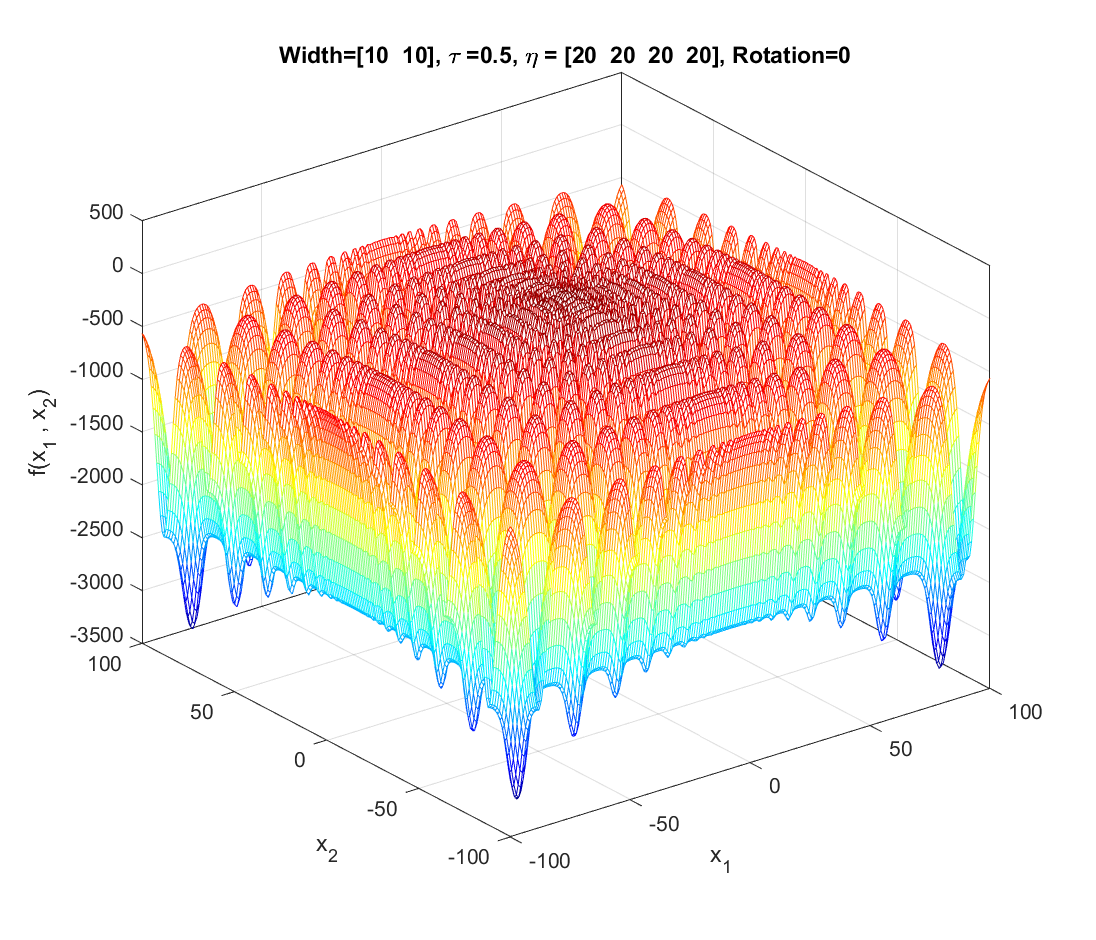}\label{fig:Cmponent:irregular(Challenging):surf}}
&
    \subfigure[]{\includegraphics[width=0.45\linewidth]{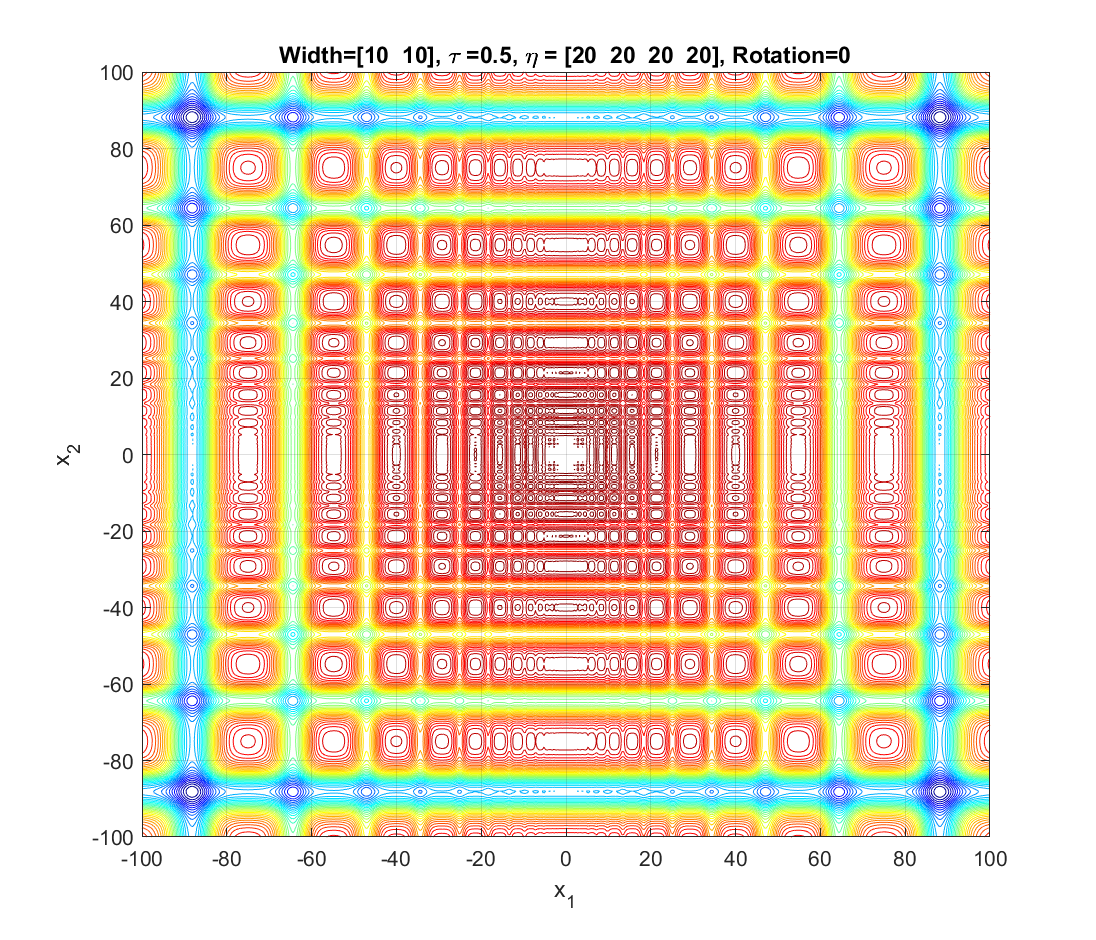}\label{fig:Cmponent:irregular(Challenging):contour}}
\end{tabular}
\caption{Three components generated by \eqref{eq:irGMPB} with different irregularity parameter settings.}
\label{fig:Cmponent:irregular}
\end{figure}

\begin{figure}[tp!]
\centering
\begin{tabular}{cc}
     \subfigure[{\scriptsize }]{\includegraphics[width=0.45\linewidth]{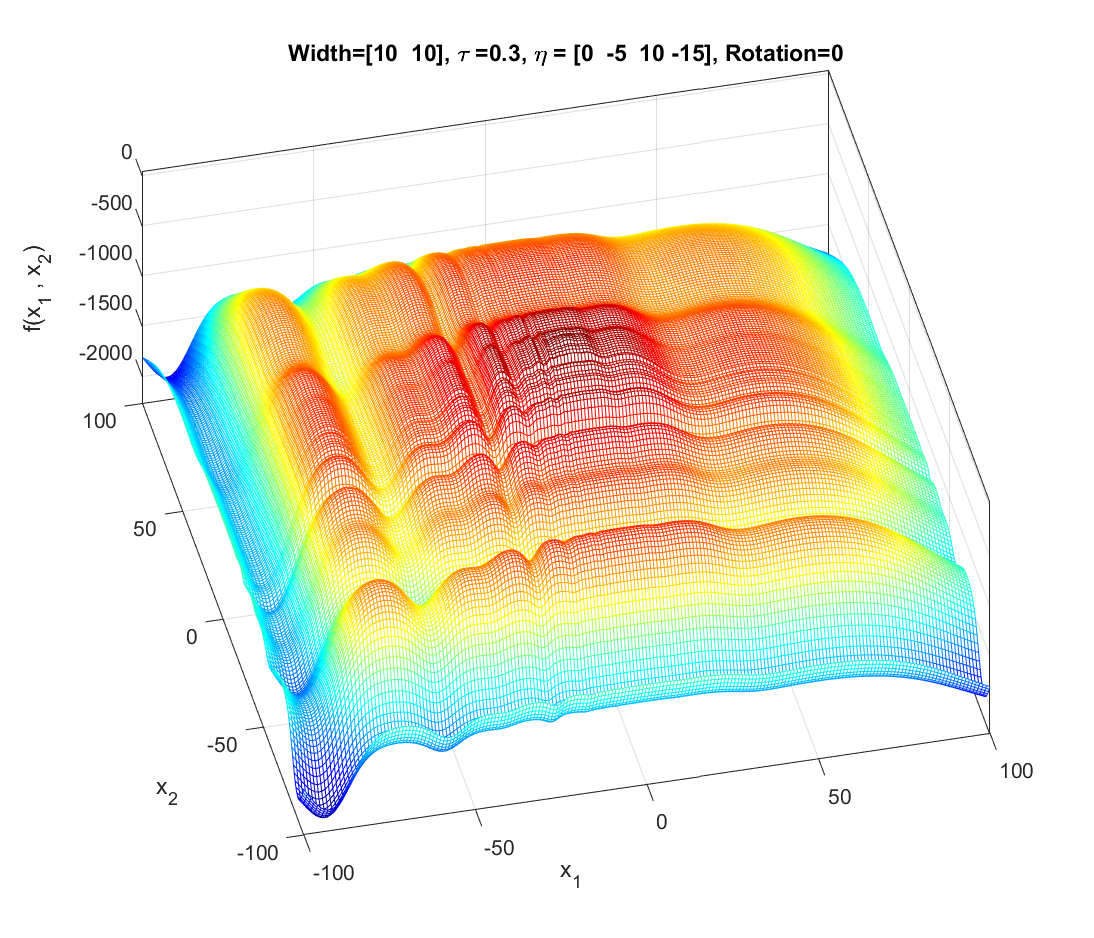}\label{fig:Component_irregular-assymetric:surf}}
&
    \subfigure[{\scriptsize }]{\includegraphics[width=0.45\linewidth]{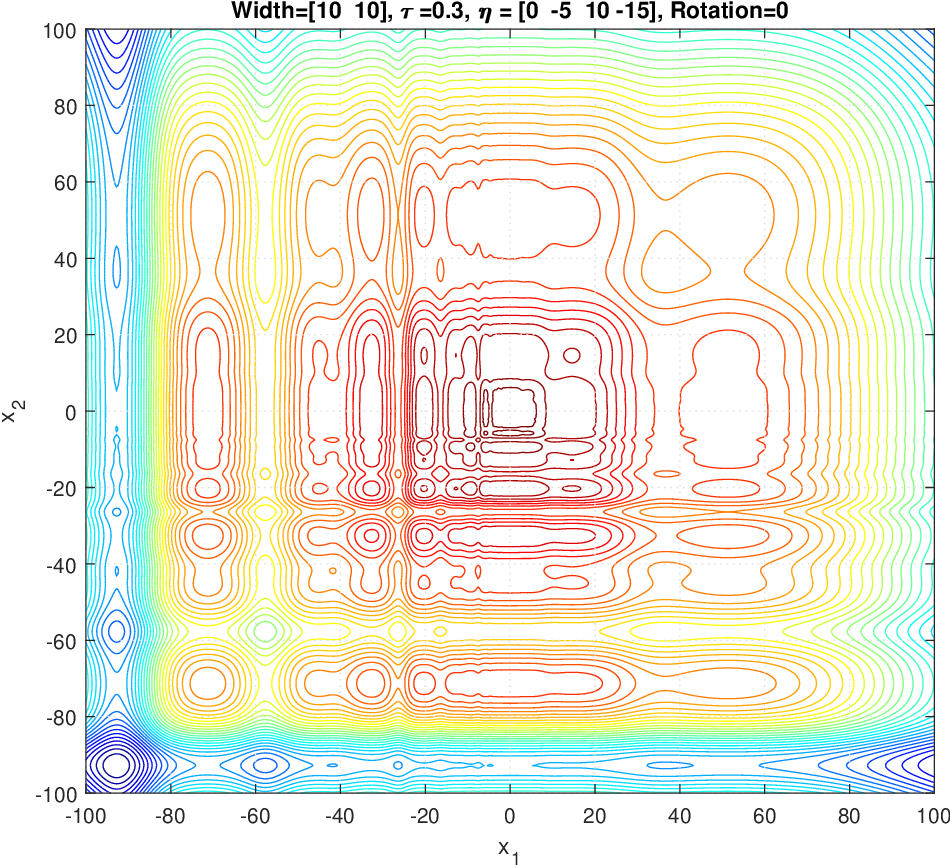}\label{fig:Component_irregular-assymetric:contour}}
\end{tabular}
\caption{An asymmetric component generated by \eqref{eq:irGMPB}.}
\label{fig:Cmponent:assymetric}
\end{figure}

In the simplest form, \eqref{eq:irGMPB} generates a symmetric, unimodal, smooth/regular, and easy-to-optimize conical peak (see Figure~\ref{fig:Cmponent:cone}).
By setting different values to  $\tau$ and $\eta$, a component generated by GMPB can become irregular and multimodal.
Figure~\ref{fig:Cmponent:irregular} shows three irregular and multimodal components generated by GMPB with different parameter settings for $\tau$ and $\eta$.
A component is symmetric when all $\eta$ values are identical, and components whose $\eta$ values are set to different values are asymmetric (see Figure~\ref{fig:Cmponent:assymetric}).

\begin{figure}[tp!]
\centering
\begin{tabular}{cc}
     \subfigure[{\scriptsize Asymmetric and ill-conditioned (condition number is two).}]{\includegraphics[width=0.45\linewidth]{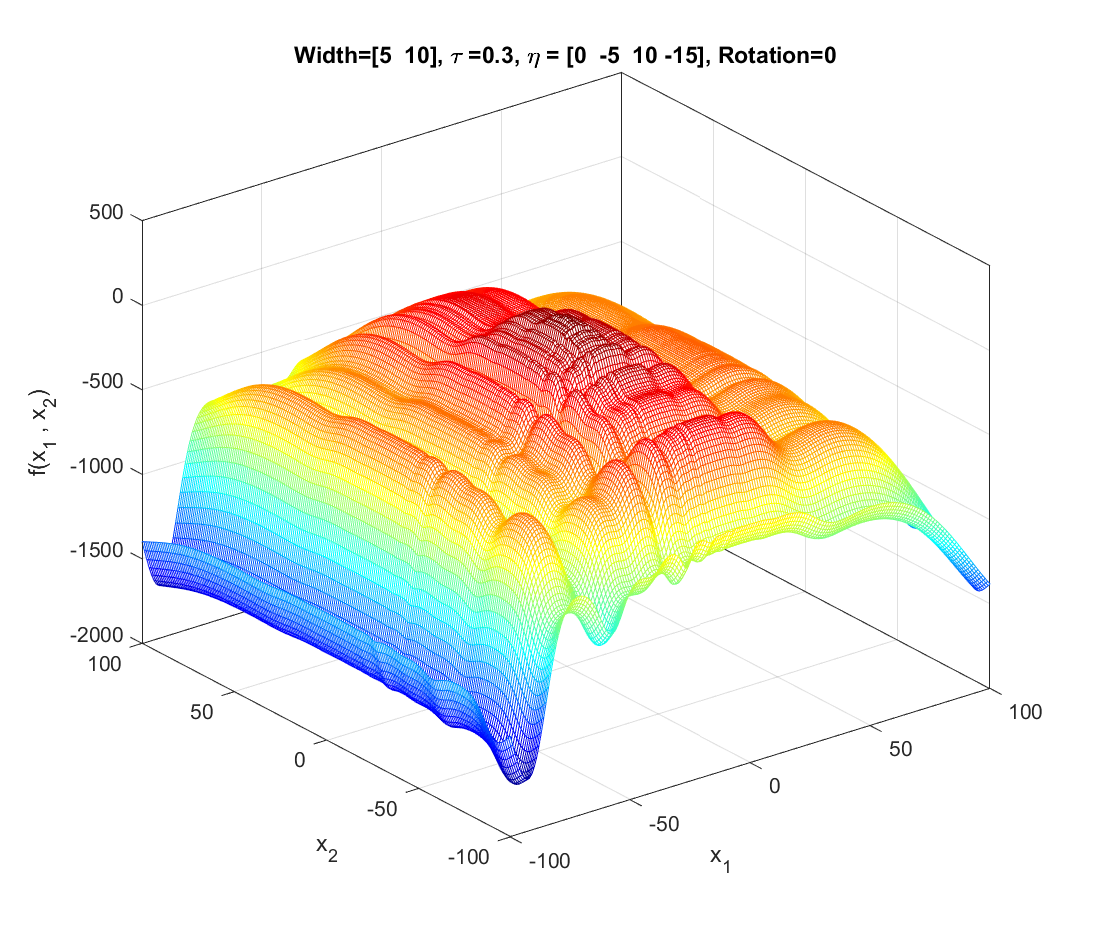}\label{fig:Component_irregular-assymetric-w77-surf}}
&
    \subfigure[{\scriptsize }]{\includegraphics[width=0.45\linewidth]{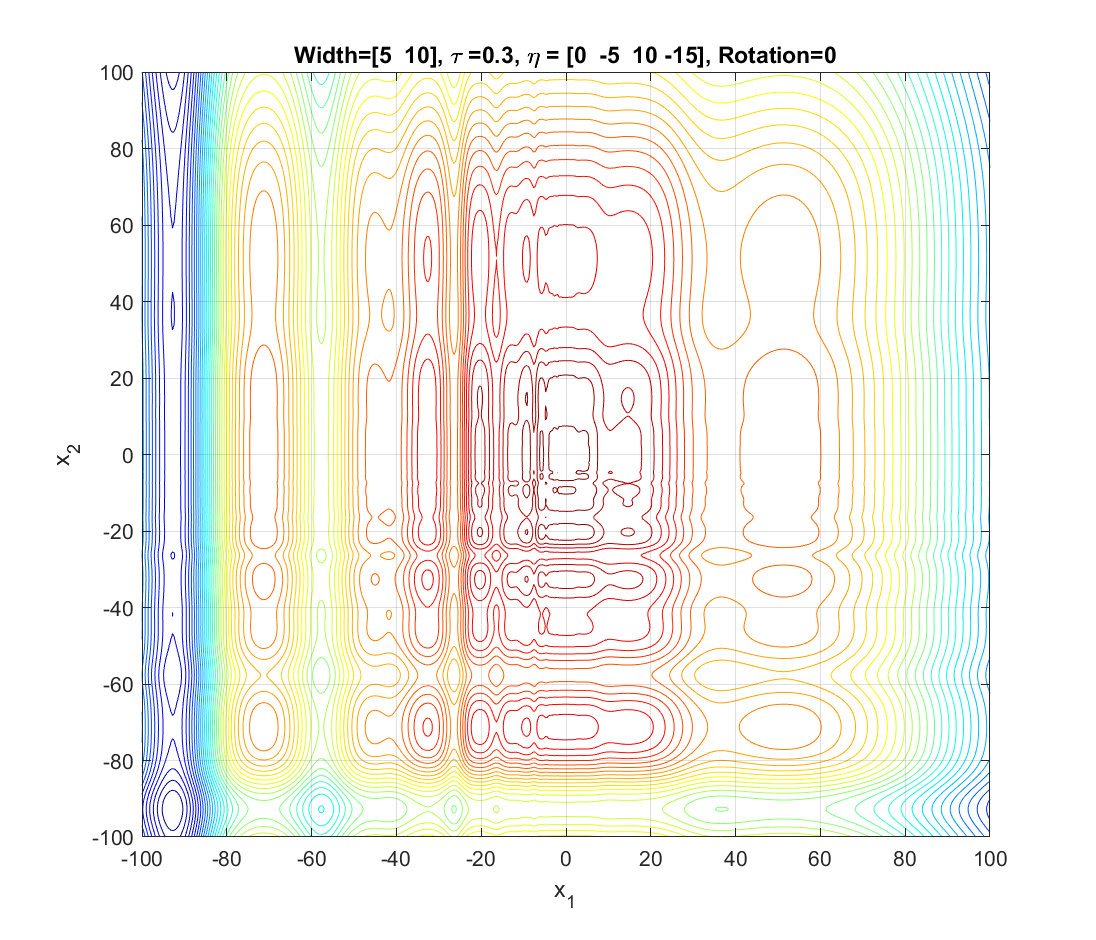}\label{fig:Component_irregular-assymetric-w77-contour}}
    \\
         \subfigure[{\scriptsize Asymmetric and ill-conditioned  (condition number is five).}]{\includegraphics[width=0.45\linewidth]{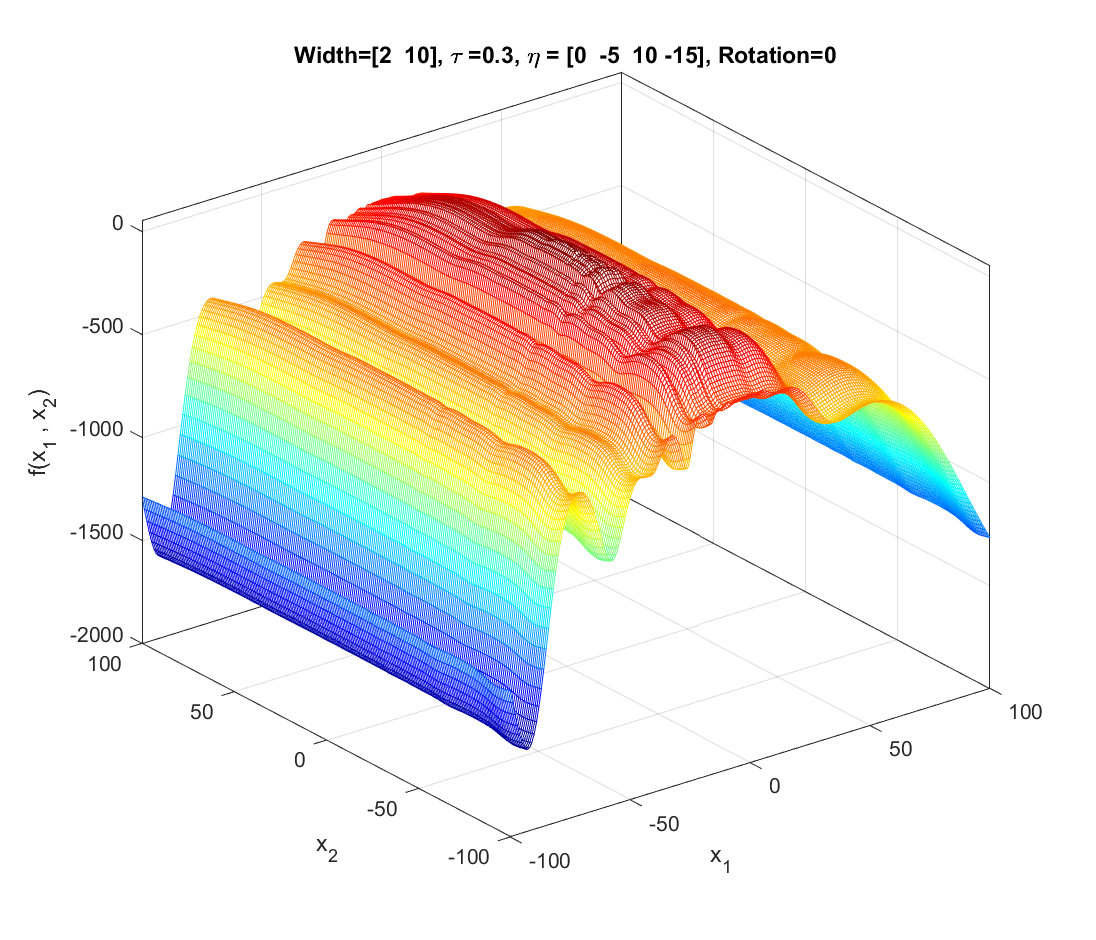}\label{fig:Component_irregular-assymetric-w73-surf}}
&
    \subfigure[{\scriptsize }]{\includegraphics[width=0.45\linewidth]{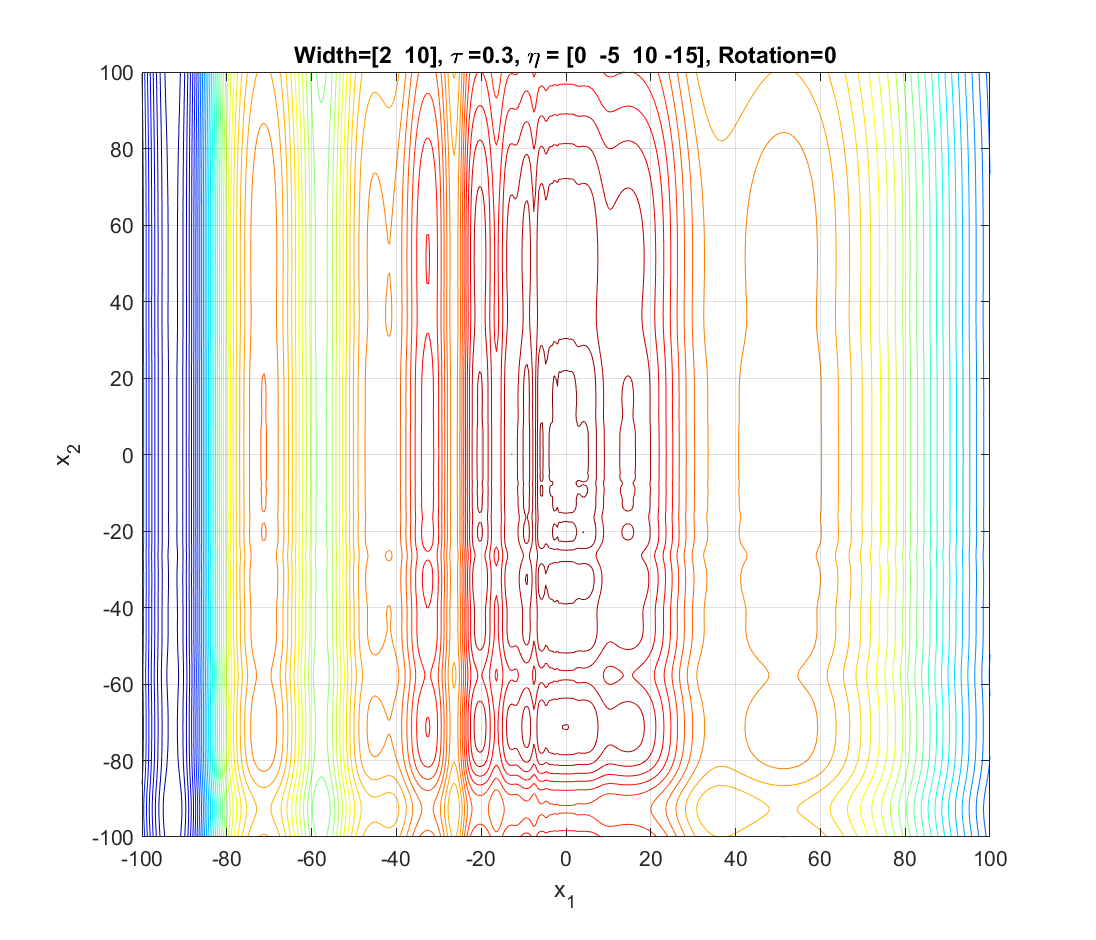}\label{fig:Component_irregular-assymetric-w73-contour}}
        \\
         \subfigure[{\scriptsize Asymmetric, ill-conditioned  (condition number is 5), and rotated.}]{\includegraphics[width=0.45\linewidth]{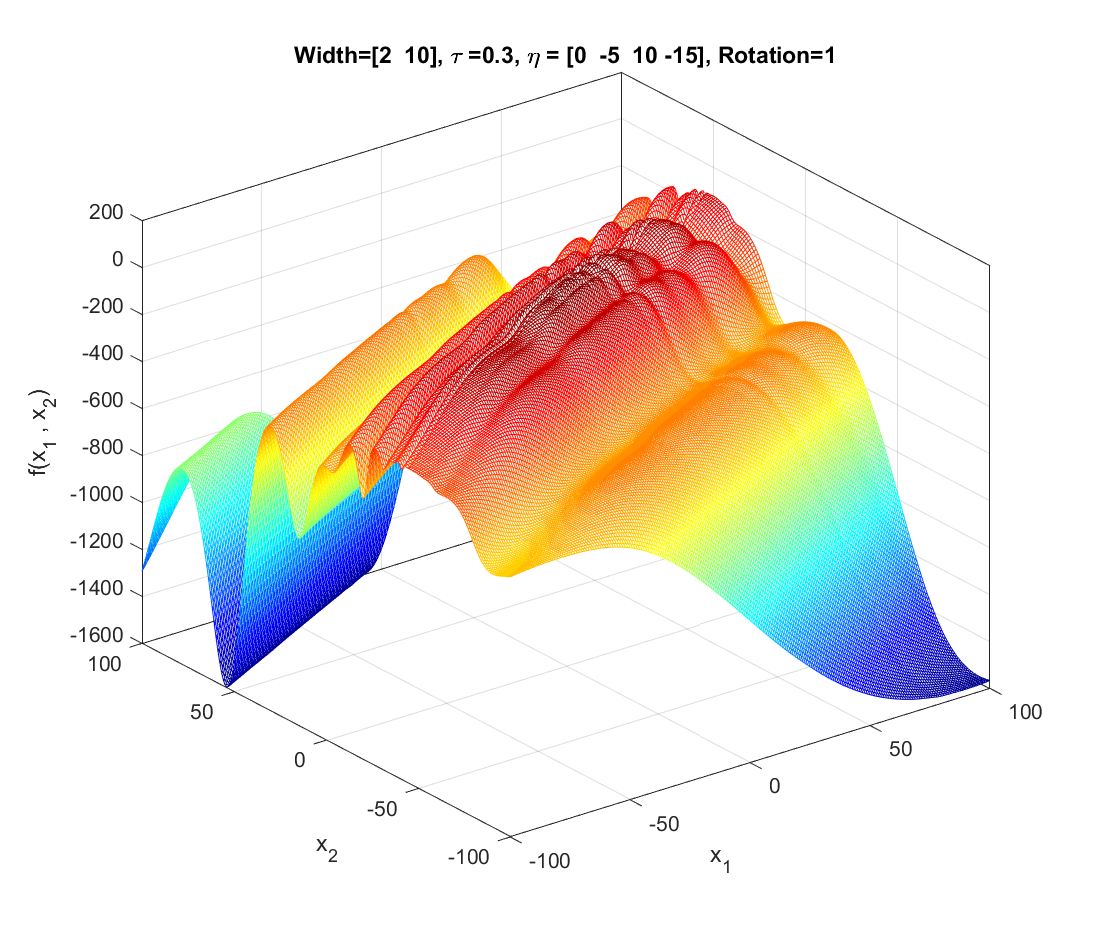}\label{fig:Rotated-surf}}
&
    \subfigure[{\scriptsize }]{\includegraphics[width=0.40\linewidth]{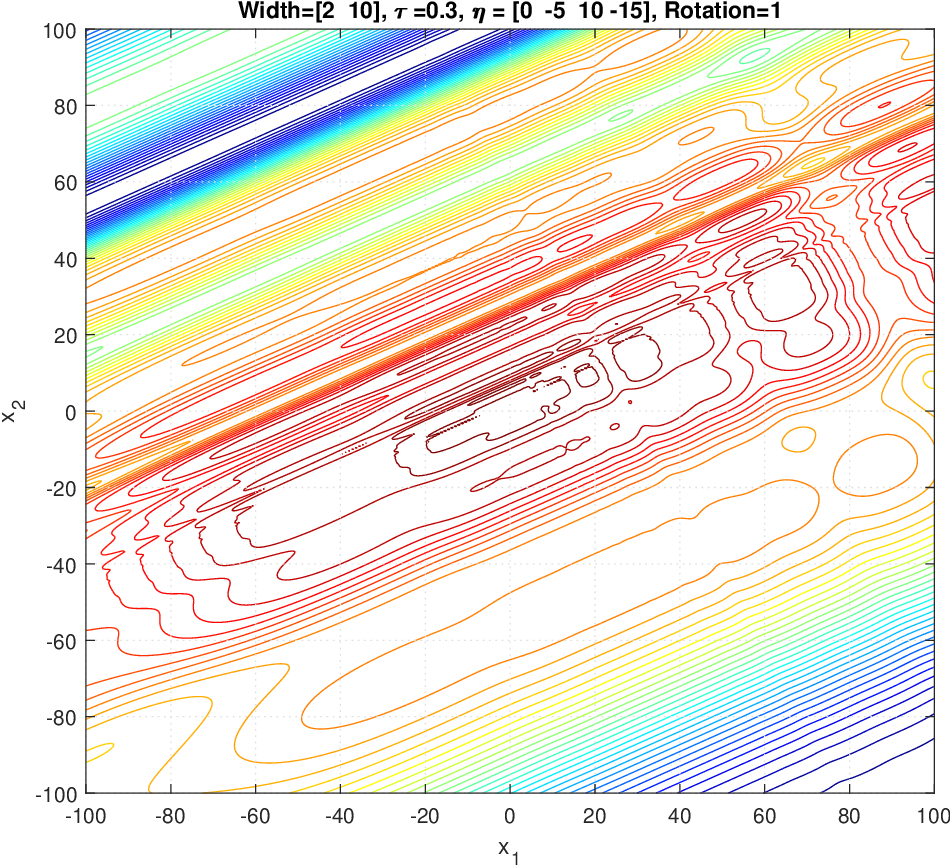}\label{fig:Rotated-contour}}
\end{tabular}
\caption{Three components generated by \eqref{eq:irGMPB} with different characteristics.}
\label{fig:Cmponent:ill-conditioning}
\end{figure}

GMPB is capable of generating components with various condition numbers.
A component generated by GMPB has a width value in each dimension.
When the width values of a component are identical in all dimensions, the condition number of the component will be one, i.e., it is not ill-conditioned.
The condition number of a component is the ratio of its largest  width value to its smallest value~\cite{yazdani2020benchmarking}. 
If a component's width value is stretched in one axis's direction more than the other axes, then, the component is ill-conditioned.
Figure~\ref{fig:Cmponent:ill-conditioning} depicts three ill-conditioned components.

In GMPB, each component $k$ is rotated using $\mathbf{R}_k$.
If $\mathbf{R}_k=\mathbf{I}$, then the component $k$ is not rotated (e.g., Figures~\ref{fig:Component_irregular-assymetric-w77-surf} and~\ref{fig:Component_irregular-assymetric-w73-surf}).
Figure~\ref{fig:Rotated-surf} is obtained by rotating the component shown in Figure~\ref{fig:Component_irregular-assymetric-w73-surf}.
In GMPB, by changing $\theta_k$ over time using \eqref{eq:angle}, the variable interaction degrees of the $k$th component change over time.


\subsubsection{Search space characteristics}
\begin{figure}[tp!]
\centering
\begin{tabular}{cc}
     \subfigure[{\scriptsize }]{\includegraphics[width=0.45\linewidth]{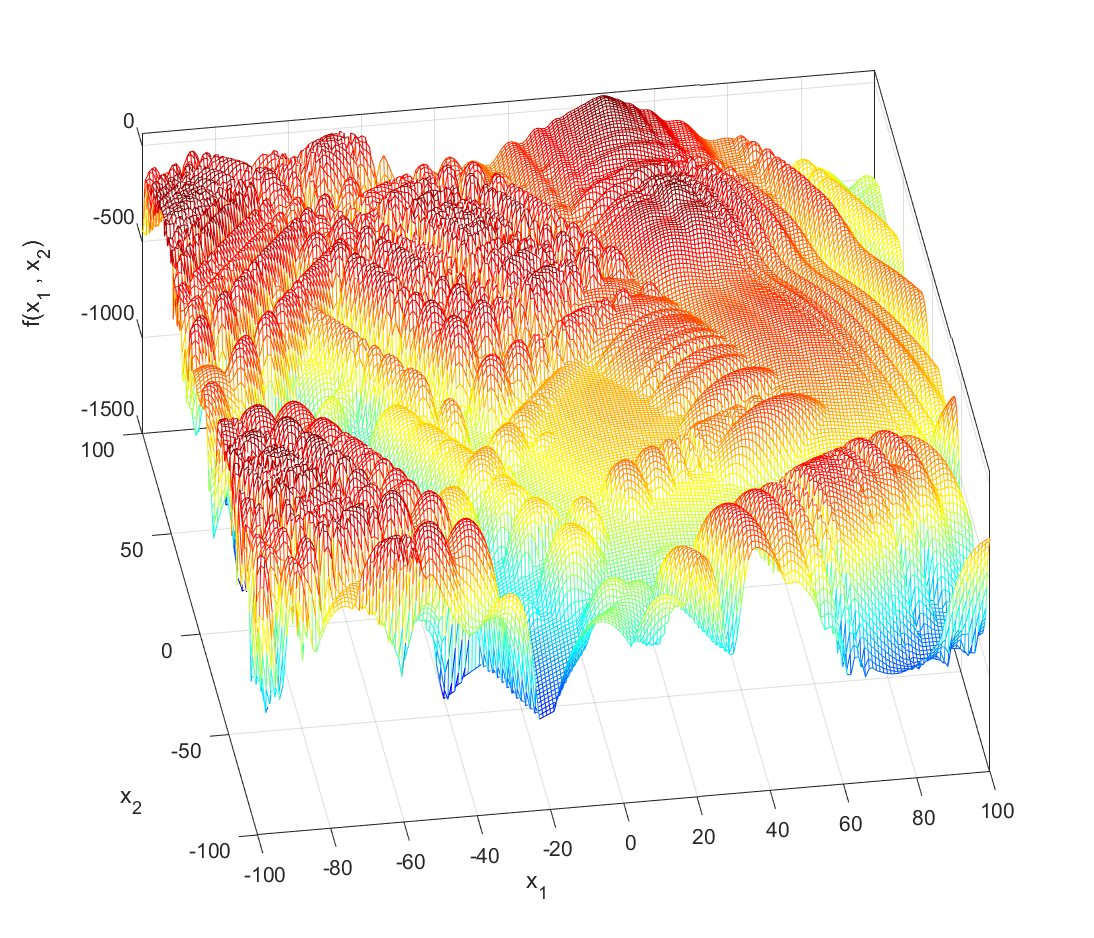}\label{fig:sub-function-surf}}
&
    \subfigure[{\scriptsize }]{\includegraphics[width=0.45\linewidth]{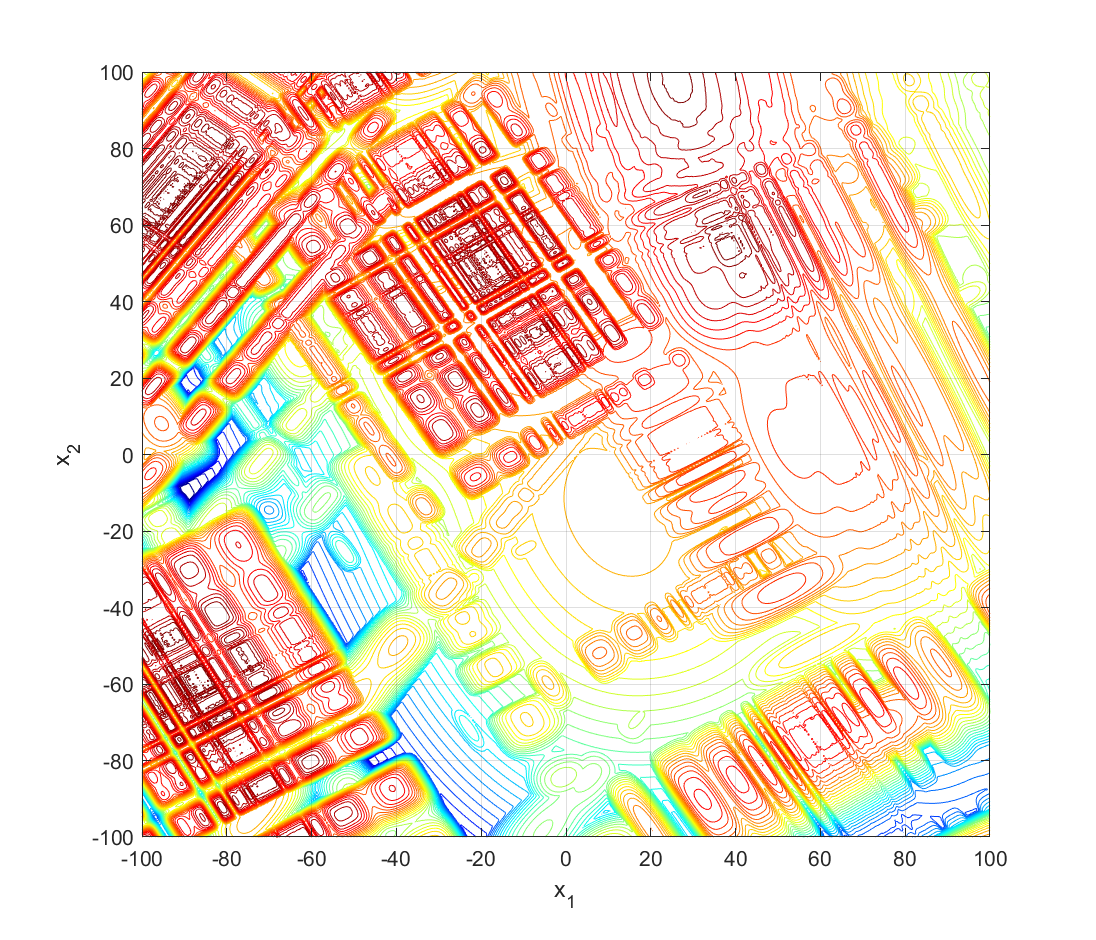}\label{fig:sub-function-contour}}
\end{tabular}
\caption{A landscape generated by \eqref{eq:irGMPB} with five components whose parameter settings are generated randomly in the ranges shown in Table~\ref{tab:GMPB-Isettings}.}
\label{fig:sub-function}
\end{figure}

Landscapes generated by GMPB are constructed by assembling several components using a $\max(\cdot)$ function in \eqref{eq:irGMPB}, which determines the basin of attraction of each component.
As shown in~\cite{yazdani2019scaling}, a landscape consists of several components (i.e., $m>1$) is fully non-separable.
Figure~\ref{fig:sub-function} shows a landscape with five components.

\section{Parameter settings}
\label{sec:senario}

%

Table~\ref{tab:GMPB-Isettings} shows the parameter settings of GMPB.
By changing the number of dimensions, shift severity value, the number of components, and change frequency, the difficulty level of the generated problem instances can be adjusted. 
Every $\vartheta$ fitness evaluations, the center position, height, width vector, angle, and irregularity parameters of each component change using the dynamics presented in~\eqref{eq:center}~to~\eqref{eq:eta}.

\begin{table*}[tp!] 
    \small
\centering
  \caption{Parameter settings of GMPB. 
  Default parameter values are highlighted where several values can be set.
  If a parameter has $k$ subscript, it belongs to the $k$th component and its value can be different from a component to another. 
  Parameters without any subscript are common between all components or they are temporal parameters. }
  \label{tab:GMPB-Isettings}
 \begin{threeparttable}
  \begin{tabular}{lll}
    \toprule
    Parameter & Symbol & Value(s)\\ 
\midrule
Dimension   & $d$	&    2,5,\hl{10},20\\
Shift severity   & $\tilde{s}$	&    \hl{1},2,5\\
Numbers of components  & $m$	             & 1,5,\hl{10},25,50,100,200\\
Angle severity     & $\tilde{\theta}$ &  $\pi/9$\\
Height severity   & $\tilde{h}$	    & 7 \\
Width severity    & $\tilde{w}$	    & 1\\
Irregularity parameter $\tau$ severity    & $\tilde{\tau}$	    &  0.2 \\
Irregularity parameter $\eta$ severity    & $\tilde{\eta}$	    &  10 \\
Search range                        &   $[Lb,Ub]^{d}$	    &    $[-100,100]^{d}$ \\
Height range                         &  $[h_{\mathrm{min}},h_\mathrm{max}]$  	    &    $[30,70]$   \\
Width range                          &   $[w_\mathrm{min},w_\mathrm{max}]^{d}$  	    &   $[1,12]^{d}$  \\
Angle range                           & $[\theta_\mathrm{min},\theta_\mathrm{max}]$ & $[-\pi,\pi]$  \\
Irregularity parameter $\tau$ range   &   $[\tau_\mathrm{min},\tau_\mathrm{max}]$ &$[0.1,1]$ \\
Irregularity parameter $\eta$ range   &  $[\eta_\mathrm{min},\eta_\mathrm{max}]$ & $[0,100]$  \\
Initial center position   & $\vec{c}^{(0)}_k$	             &  $\mathcal{U}[Lb,Ub]^{d}$\\
Initial height                 &   $h^{(0)}_k$       &  $\mathcal{U}[h_{\mathrm{min}},h_\mathrm{max}]$ \\
Initial width                          & $\vec{w}^{(0)}_k$ & $\mathcal{U}[w_\mathrm{min},w_\mathrm{max}]^{d}$  \\
Initial angle                            & $\theta^{(0)}_k$ & $\mathcal{U}[\theta_\mathrm{min},\theta_\mathrm{max}]$ \\
Initial irregularity parameter $\tau$    & $\tau^{(0)}_k$    &  $\mathcal{U}[\tau_\mathrm{min},\tau_\mathrm{max}]$  \\
Initial irregularity parameter $\eta$    &$\eta^{(0)}_k$    &  $\mathcal{U}[\eta_\mathrm{min},\eta_\mathrm{max}]$   \\    
Initial rotation matrix    &$\mathbf{R}^{(0)}_k$    &   $\mathrm{GS}(\mathcal{N}(0,1)^{d\times d})$\tnote{\textbf{$\dagger$}}  \\   
Change frequency   & $\vartheta$	             & 1000,2500,\hl{5000},10000\\
Number of Environments                 &   $T$       &  100 \\
    \bottomrule
  \end{tabular}
  \begin{tablenotes}
   \begin{scriptsize}
  \item[$\dagger$] $\mathbf{R}_k$ is initialized by performing the Gram-Schmidt orthogonalization method $\mathrm{GS}(\cdot)$ on a $d \times d$ matrix with normally distributed entries.
   \end{scriptsize}
    \end{tablenotes}
 \end{threeparttable}
 \end{table*}

\section{Performance indicator}
\label{sec:PerformanceIndicator}

To measure the performance of algorithms in solving the problem instances generated by GMPB, the \emph{offline-error}~\cite{branke2003designing} ($E_{\mathrm{O}}$) and the average error of the best found solution in all environments, i.e., the best error before change, ($E_\mathrm{BBC}$)~\cite{trojanowski1999searching} are used as the performance indicators. 
$E_{\mathrm{O}}$ is the most commonly used performance indicator in the literature~\cite{yazdani2021DOPsurveyPartB}. 
This approach calculates the average error of the best found position over all fitness evaluations using the following equation:
\begin{align}
\label{eq:Eo}
E_{\mathrm{O}} = \frac{1} { T\vartheta } \sum_{t=1}^T \sum_{\varphi=1}^\vartheta \left(f^{(t)}\left( \vec{x}^{\star(t)} \right) - f^{(t)}\left(\vec{x}^{*((t-1)\vartheta + \varphi)}\right)\right),
\end{align}
where $\vec{x}^{\star(t)}$ is the global optimum position at the $t$th environment, $T$ is  the number of environments, $\vartheta$ is the change frequency, $\varphi$ is the fitness evaluation counter for each environment, and $\vec{x}^{*((t-1)\vartheta+\varphi)}$ is the best found position at the $\varphi$th fitness evaluation in the $t$th environment.

$E_\mathrm{BBC}$ is the second commonly used performance indicator in the field~\cite{yazdani2021DOPsurveyPartB}.
 $E_\mathrm{BBC}$ only considers the last error before each environmental change (i.e., at the end of each environment): 
\begin{align}
\label{eq:Eb}
E_\mathrm{BBC} = \frac{1}{T} \sum_{t=1}^T  \left(f^{(t)}\left( \vec{x}^{\star(t)} \right) - f^{(t)}\left(\vec{x}^{*(t)}\right)\right),
\end{align}
where $\vec{x}^{*(t)}$ is the best found position in $t$th environment which is fetched at the end of the $t$th environment.

\section{Source code}

GMPB is a featured component of EDOLAB~\cite{peng2023evolutionary}, a comprehensive MATLAB\footnote{Version R2020b and later} optimization platform designed for both educational and experimental purposes in dynamic environments. 
EDOLAB hosts an array of 25 evolutionary dynamic optimization algorithms. 
Researchers are encouraged to utilize the guidelines in~\cite{peng2023evolutionary} to integrate their own algorithms into EDOLAB.
This allows for an extensive evaluation of their algorithms' performance against a variety of problem instances created by dynamic optimization benchmark generators, including GMPB. 
The source code of EDOLAB is readily available and can be accessed at \href{https://github.com/EDOLAB-platform/EDOLAB-MATLAB}{\textcolor{blue}{\underline{here}}}.

We also provided another source code that visualizes 2-dimensional landscapes generated by GMPB which can be found in~\cite{yazdani2021GMPBvisualization}.
Using this source code,  researchers can observe the impacts of different parameter settings on morphological characteristics of the problems generated by GMPB.

\bibliography{bib} 
\bibliographystyle{IEEEtran}

\end{document}